\documentclass[letterpaper, 10 pt, conference]{ieeeconf}  

\IEEEoverridecommandlockouts                              
\usepackage[utf8]{inputenc}
\usepackage[T1]{fontenc}

\usepackage{amsmath,amssymb,amsfonts}

\usepackage{amsthm}
\usepackage{setspace}
\usepackage{xcolor}
\usepackage{graphicx}   
\usepackage[ruled,vlined,linesnumbered]{algorithm2e}
\usepackage[caption=false]{subfig}
\usepackage{cite}

\newtheorem{defn}{Definition}
\newtheorem{exmp}{Example}
\newtheorem{rem}{Remark}
\newtheorem{prob}{Problem}

\DeclareMathOperator*{\argmin}{\arg\min}

\newcommand{\Real}{\mathbb{R}}
\newcommand{\Bool}{\mathbb{B}}

\newcommand{\Ss}{\mathcal{X}}

\title{\LARGE \bf
Risk-Aware Motion Planning in Partially Known Environments
}

\author{Fernando S. Barbosa$^{1}$, Bruno Lacerda$^{2}$, Paul Duckworth$^{2}$, Jana Tumova$^{1}$ and Nick Hawes$^{2}$
\thanks{*This work was partially supported by Sweden’s  Innovation  Agency  (Vinnova) through the TECoSA project, UK Research and Innovation and EPSRC through the Robotics and Artificial Intelligence for Nuclear (RAIN) hub [EP/R026084/1], the EPSRC Programme Grant `From Sensing to Collaboration' [EP/V000748/1], and a gift from Amazon Web Services.}
\thanks{$^{1}$KTH Royal Institute of Technology and Digital Futures, Stockholm, Sweden.
        {\tt\small \{fdsb, tumova\}@kth.se}}%
\thanks{$^{2}$Oxford Robotics Institute, University of Oxford, United Kingdom.
        {\tt\small \{bruno, pduckworth, nickh\}@robots.ox.ac.uk}}%
}
\begin{document}

\maketitle
\thispagestyle{empty}
\pagestyle{empty}

\begin{abstract}
Recent trends envisage robots being deployed in areas deemed dangerous to humans, such as buildings with gas and radiation leaks.
In such situations, the model of the underlying hazardous process might be unknown to the agent \textit{a~priori}, giving rise to the problem of planning for safe behaviour in partially known environments.
We employ Gaussian process regression to create a probabilistic model of the hazardous process from local noisy samples. The result of this regression is then used by a risk metric, such as the Conditional Value-at-Risk, to reason about the safety at a certain state. 
The outcome is a risk function that can be employed in optimal motion planning problems.
We demonstrate the use of the proposed function in two approaches.
First is a sampling-based motion planning algorithm with an event-based trigger for online replanning.
Second is an adaptation to the incremental Gaussian Process motion planner (iGPMP2), allowing it to quickly react and adapt to the environment.
Both algorithms are evaluated in  representative simulation scenarios, where they demonstrate the ability of avoiding high-risk areas. 
\end{abstract}

\section{INTRODUCTION}
One of the appeals of robots is that they can be deployed in environments that are dangerous to humans. Such applications include search-and-rescue in disaster scenarios, identification of gas leaks, and inspection of radioactive facilities. In most of these domains, a model of the underlying hazardous process (temperature, gas concentration, radiation activity) cannot be provided to the robot prior to execution, thus observations of the process must be made via (noisy) sensor readings, and a model built on demand. Since exposing a robot to certain amounts of such hazards may damage it, operation in such conditions requires reasoning about the dangers unexplored areas might bring. This gives rise to the problem of risk-aware motion planning in partially-known and partially-observable environments; partially-known due to the lack of knowledge about the distribution of the hazardous process on the environment, and partially-observable due to noisy sensor readings of a process.

In this paper, we propose a method to tackle  the challenge of finding a trajectory that reaches a goal region while minimizing the risk of passing through hazardous regions, \textit{without} a prior model of the hazard.
We use a Gaussian process (GP) to model an \emph{a priori} unknown process from noisy samples taken along the trajectory already traversed by the robot. 
We then exploit the fact that the posterior of a GP evaluated at an unvisited state is a probability distribution, and use it to construct a risk function. The distribution represents the values of the hazard that the robot could experience by visiting that state. A risk metric, such as the Conditional Value at Risk (CVaR), over such a distribution returns a conservative expected value that accounts for less-likely tail events.
Given user-defined constraints on the acceptable values of the hazard, we use the posterior as input to a risk metric, and transform that into a risk function to be employed in optimal motion planning problems. 

The contribution of this paper is twofold. One contribution is the proposal of a risk-aware cost function, which converts the risk measures over a GP posterior into a cost function for planning. The novelty lies on the process of transforming the probability distribution returned by the GP regression into a deterministic value, which in turn accounts for tail-events of such a distribution. Furthermore, we also propose a conservative gradient of the risk-function, which can be used by optimization-based planners to speed up convergence.
The second contribution is the demonstration of how such a risk function can be used in two online motion planning algorithms. 
First we describe a sampling-based planning algorithm with an event-triggered replanning, in which the optimal trajectory is a balance between the trajectory length and the risk of violating the constraint on the unknown feature.
Second, we show how the proposed risk function can be used in an optimization-based planning algorithm, which accounts not only for trajectory length, but also smoothness and distance to obstacles.
Simulation results are presented for both algorithms.

\section{RELATED WORK}
Risk-aware motion planning algorithms are usually developed to minimize the probability of collisions. Collisions might occur due to imprecise localization of the robot and obstacles, inaccurate models of the dynamics of the robot itself or of the dynamic obstacles, or even due to disturbances \cite{luders2013robust, hakobyan2019risk, chi2017risk}. In our work we employ risk metrics, such as CVaR \cite{majumdar2020should}, in order to take into consideration less likely - but probably dangerous - tail events of the distribution returned by GP regression, and use that to construct a risk-aware cost function. \cite{suh2012cost, devaurs2015optimal} provide approaches to motion planning on complex cost spaces, and \cite{suresh2019planning} to uncertain cost spaces. Neither of them, however, deal with a dynamic cost function.

GPs have been widely used in the literature for modelling spatial and spatiotemporal processes, as well as for learning models of dynamical systems. A periodic sampling approach on a discretized grid over the environment is proposed in \cite{carron2016machine}, in which the authors explore links between GPs and Kalman filters. Instead of sampling on a fixed grid,  \cite{marchant2014sequential} proposes a sequential sampling approach for an agent navigating in the environment. The motivation behind it is that, in robotic applications, not only where and when samples are collected are important, but also at which order they are taken. \cite{flaspohler2019information} proposes an information-guided approach to finding the maximum value of a partially-observable distribution. \cite{duckworth2020time} plan over a discrete and time-varying domain using a GP to inform a Monte Carlo tree search. However, \cite{marchant2014sequential, flaspohler2019information, duckworth2020time} do not consider cases where the underlying process might pose a safety risk to the agent, as we do.

Safe exploration of spatial processes is approached in~\cite{turchetta2016safe,buddmarkov} with the use of Markov decision processes with unknown state features values. The goal of \cite{buddmarkov} is to converge to a good approximation of the process without violating a safety threshold. In interactive machine learning, \cite{turchetta2019safe} proposes a framework that renders any unsafe learning algorithm safe by exploring regularity assumptions of GP priors. Neither of these approaches attempts to reach a target location safely.
Safety is also taken into consideration when dealing with dynamical systems, as in \cite{koller2018learning} and \cite{jackson2020safety}. In the former the authors design a model predictive control approach that provides high probability safety guarantees while learning the dynamics of the system, while in the latter a safety verification approach is proposed.
In contrast our work does not aim to explore the space in order to converge to a precise model of the \emph{a priori} unknown process or dynamics. Our problem is that of navigating from an initial to a goal state in the environment while minimizing the risk of being exposed to undesired values of a spatial process. A recent review of sampling-based motion planning algorithms \cite{gammell4survey} shows that this is still an open problem.

Gaussian Processes have also been used for optimal motion planning. In \cite{mukadam2018continuous}, continuous trajectories are represented by sparse GP models, and three algorithms for planning such trajectories are proposed, GPMP, GPMP2 and iGPMP2. We show how our risk-based cost function can be used in these algorithms, with special attention to iGPMP2, which is an incremental algorithm that allows for fast online replanning.

In previous works \cite{barbosa2019guiding,karlsson2020sampling} we proposed using soft spatial constraints given by signal temporal logic (STL) specifications to find trajectories that minimally violate user constraints. In both works STL predicates were defined purely as functions of the trajectory and its relation with the fully known, deterministic environment, such as the distance to obstacles and WiFi routers. In this work we define similar constraints over a partially-observable environment, which is unknown \emph{a priori} and approximated by a GP over noisy and incremental observations. The threshold constraint we use corresponds to a fragment of STL, and we leave the extension to more general classes of STL to future work.

\section{Preliminaries} \label{sec:Preliminaries}
Booleans are denoted by $\Bool$, real numbers by $\Real$ and the set of all nonnegative reals by $\Real_{\ge 0}$, while $\Real^n$ is the $n$-dimensional real vector space. Let $\Ss \subset \Real^n$ be a bounded environment divided into obstacle- and free-space, respectively $\Ss_{\text{obs}} \subset \Ss$ and $\Ss_{\text{free}} \subseteq \Ss$, with $\Ss_{\text{free}}~:=~\Ss \setminus \Ss_{\text{obs}}$. 

\begin{defn}[Trajectory]\label{def:traj}
 A trajectory between two states $x_1, x_2 \in \Ss$ is a piecewise smooth curve $C \subset \Ss$ with a bijective parametrization $\varsigma: [0,1] \to C$ such that $\varsigma(0)~=~x_1$ and $\varsigma(1) = x_2$. The set of all trajectories is denoted by $\Sigma$.
\end{defn}

\begin{defn}[Unknown process]
Let $g: \Ss \to \Real$ be an \emph{a priori} unknown process from which the agent can draw noisy sensor measurements $z:~\Ss \to \Real$ according to the following observation model 
\begin{equation}
    z(x) = g(x) + \epsilon,
\end{equation}
where $\epsilon \sim \mathcal{N}(0,\sigma_n^2)$ is zero-mean additive sensor noise.
\end{defn}

\begin{defn}[Optimal Motion Planning problem]
Given a cost functional $J: \Sigma \to \Real_{\ge 0}$, an initial state $x_{\text{init}} \in \Ss$ and a goal region $\Ss_{\text{goal}} \subset \Ss$, the optimal solution is a trajectory~$\varsigma^*$ that minimizes the cost functional, i.e.
\begin{align}
    \varsigma^* = \argmin_{\varsigma \in \Sigma} J(\varsigma),
\end{align}
while satisfying the following constraints: $\varsigma^*(0) = x_{\text{init}}$, $\varsigma^*(1) \in \Ss_{\text{goal}}$, and $\varsigma^*(t) \in \Ss_{\text{free}}$, $\forall t \in [0,1]$.
\end{defn}

The cost functional $J$ can encode a metric over the quality of a trajectory, which can include, for instance, trajectory length and smoothness, as well as distance to obstacles.

\subsection{Gaussian Processes (GP)}\label{sec:GP}
Let the belief over the process $g$ be modeled as a zero-mean GP with covariance (kernel) $\kappa : \Ss \times \Ss \to \Real$. A data set of $N$ (noisy) sensor measurements is given by $\mathcal{D}~=~\{x_i,z(x_i)\}^N_{i=1}$. 

\begin{defn}[GP posterior]\label{def:gp}
\cite{rasmussen2003gaussian} The posterior belief $\hat{g}$ at a position $x'$ is computed by $\hat{g}(x') \mid \mathcal{D} \sim \mathcal{N}(\mu(x'), \sigma^2(x'))$ where
\begin{align}
    \mu(x') &= \boldsymbol{\kappa}(x') \mathbf{\bar{K}}^{-1} \mathbf{z}, \\
    \sigma^2(x') &= \kappa(x',x') - \boldsymbol{\kappa}(x') \mathbf{\bar{K}}^{-1} \boldsymbol{\kappa}^T(x'),
\end{align}
with $\mathbf{z} = [z_1, \dots, z_N]^T$, $\mathbf{\bar{K}} = \bar{K} + \sigma_n^2 I$ where $\bar{K} \in \Real^{N \times N}$ a positive definite matrix with $\bar{K}[i,j] = \kappa(x_i,x_j)$, and $\boldsymbol{\kappa}(x') = [\kappa(x_1,x'), \dots, \kappa(x_N,x')]$.
\end{defn}

Defining an appropriate kernel $\kappa$ and its hyperparameters is problem specific. We direct the reader to \cite{rasmussen2003gaussian} for a more complete material on GPs.

Furthermore, it holds that the derivative of a GP is also a GP \cite{rasmussen2003gaussian}. Therefore, the posterior belief of the derivative of~$\hat{g}$ is given by $\frac{\partial\hat{g}(x')}{\partial x} \mid \mathcal{D} \sim \mathcal{N}(\frac{\partial\mu(x')}{\partial x}, \frac{\partial\Sigma(x')}{\partial x})$ with
\begin{align}
\small
    \frac{\partial\mu(x')}{\partial x} &= \left.\frac{\partial\boldsymbol{\kappa}(x)}{\partial x}\right|_{x'} \mathbf{\bar{K}}^{-1} \mathbf{z}, \\
    \frac{\partial\Sigma(x')}{\partial x} &= \left.\frac{\partial^2\kappa(x,x)}{\partial x \partial x}\right|_{x'} - \left.\frac{\partial\boldsymbol{\kappa}(x)}{\partial x}\right|_{x'} \mathbf{\bar{K}}^{-1} \left(\left.\frac{\partial\boldsymbol{\kappa}(x)}{\partial x}\right|_{x'}\right)^T.
\end{align}

\subsection{Risk metrics}
It is important that safety-critical systems present a certain level of robustness over probabilistic outcomes. Simply employing the expected value operator $\mathbb{E}[Z]$ over a random variable $Z$ does not reason over the possible outcomes in the tails of the distribution, while using the worst-case value is too conservative.
A risk metric allows an agent to calculate the perceived risk of a random variable $Z$. We use two existing risk metrics, VaR and CVaR, and direct the reader to~\cite{majumdar2020should} for a description of risk metrics for robotics applications.

\begin{defn}[Risk metric]\label{def:risk}
\cite{majumdar2020should} Let $(\Omega, \mathcal{F}, \mathbb{P})$ denote a probability space with $\Omega$ the sample space, $\mathcal{F}$ the event space, and $\mathbb{P}$ the probability function $\mathbb{P}: \mathcal{F} \to \Real$, and let $Z: \Omega \to \Real$ be a random variable. The set of all random variables $Z$ is denoted by $\mathcal{Z}$. A risk metric maps a random variable to a real number, i.e. it is a mapping $r: \mathcal{Z} \to \Real$.
\end{defn}

\begin{defn}
The Value at Risk (VaR) of a random variable~$Z$ at level $\beta$ is the $(1-\beta)$-quantile of $Z$, i.e. 
\begin{equation}
    \text{VaR}_{\beta}(Z) = min\{z \mid \mathbb{P}(Z>z) \le \beta\}.
\end{equation}
\end{defn}

\begin{defn}
The Conditional Value at Risk (CVaR) of $Z$ at level $\beta$ is the expected value of $Z$ in the conditional distribution of $Z$'s upper $(1-\beta)$-tail 
\begin{equation}
    \text{CVaR}_{\beta}(Z) = \mathbb{E}[Z' \mid Z' > \text{VaR}_{\beta}(Z)].
\end{equation}
\end{defn}

Similarly, VaR and CVaR can be defined to consider the lower tail instead.

\section{Problem Formulation and Solution}
The problem targeted by this work is that of optimal motion planning in partially known environments. This partial knowledge is due to the presence of an \emph{a priori} unknown process $g$, which might pose a danger to the agent's operation; we assume $\Ss_{\text{obs}}$ to be known. Loosely speaking, our problem is to find a trajectory that balances its length with the level of violation of a user-defined constraint over the unknown process.

Let us connect the risk metric $r$ (Def.~\ref{def:risk}) with a GP~$\hat{g}$ (Def.~\ref{def:gp}) and a trajectory~$\varsigma$ (Def.~\ref{def:traj}) by saying that the risk value at a safety level $\beta$ of a state at time $t$ along a trajectory~$\varsigma$ is defined as $r_\beta(\hat{g}(\varsigma(t)))$, for $t \in [0,1]$. One can now use this to define a desired upper-bound $\alpha \in \Real$ on the risk value by defining a constraint\footnote{Note that $\varphi$ resembles an STL predicate. We plan to explore the link to more complex STL formulas in future work.} $$\varphi = \alpha - r_\beta(\hat{g}(\varsigma(t))).$$ A constraint $\varphi$ is being violated at a time $t$~if~$\varphi(t)~<~0$.

\begin{prob}
Given an unknown process $g$ on an environment~$\Ss$, find a trajectory $\varsigma$ from an initial state $x_{\text{init}}~\in~\Ss_{\text{free}}$ to a goal region $\Ss_{\text{goal}} \subset \Ss_{\text{free}}$ that avoids collisions with obstacles $\Ss_{\text{obs}}$, and minimally violates the risk-aware constraint~$\varphi$.
\label{prob:prob1}
\end{prob}

\begin{exmp}
    Let us suppose we have an agent that is to be deployed into a building with a gas leak in order to reach an office and check for survivors. The agent can operate on high concentrations of gas, but it is best staying below a certain threshold $\alpha$. Thus we can write the constraint $\varphi$ as $\varphi = \alpha - \text{CVaR}_{0.05}(\hat{g}(\varsigma(t)))$, where we choose the risk metric to be the CVaR with safety level of 95\%.
\end{exmp}

To solve Problem~\ref{prob:prob1} we rely on iteratively constructing a model based on a data set $\mathcal{D}$ of local samples of $g$. This has two implications. First, we require an online motion planning algorithm, since the robot has to navigate towards its goal in order to collect more samples of $g$. Second, we require a way of penalizing trajectories that traverse areas that could possibly violate the safety specifications. For such, we propose a risk cost function, described next, and employ it in two motion planning algorithms, described in Sec.~\ref{sec:samplingApproach}-\ref{sec:gpmp}.

\subsection{Risk cost function}
The posterior belief $\hat{g}$ at a state $x' \in \Ss$, given a data set $\mathcal{D}$ and an appropriate kernel $\kappa$, is a normal distribution with mean $\mu(x')$ and covariance $\sigma^2(x')$ estimated via GP regression (Sec.~\ref{sec:GP}). A risk function $r$ can use $\hat{g}(x')$ as its random variable and estimate the (conditional) value-at-risk at safety levels $\beta$.

We propose a risk cost function $f:~\Ss~\to~\Real$ to assign costs to states according to the level of violation of $\varphi$. Given a constraint $\varphi$, we propose $f$ to be defined as follows, with $\gamma > 0$ a user-defined parameter
\begin{equation}
    f(x) = \max(e^{-\gamma\varphi(x)}, 1).
    \label{eq:field}
\end{equation}

Note that $\varphi(x) < 0$ only when the constraint $\varphi$ is being violated; in such situation, $f > 1$. Whenever $\varphi$ is being satisfied at state $x$, $f(x) = 1$.

Some optimization-based planners take advantage of the gradient of the cost function in order to speed-up convergence to a local minimum. Denoting the gradient of $f$ in relation to $x$ as $\frac{d f(x)}{dx}$, we derive the following:
\begin{equation}\label{eq:gradient}
    \frac{d f(x)}{dx} = \begin{cases} 
        0, \text{ if }\varphi(x) \ge 0 \\[1ex]
        \gamma r_\beta\left(\dfrac{d\hat{g}(x)}{dx}\right) e^{-\gamma\varphi(x)} \text{, otherwise.}
        \end{cases}
\end{equation}

For the last step we assume that
$\dfrac{dr_\beta(\hat{g}(x))}{dx}~=~r_\beta\left(\dfrac{d\hat{g}(x)}{dx}\right)$. We argue that this is a quite fair assumption as we recall that a risk metric returns a (probably) conservative expectation of the probability distribution. If no risk metric were being used, \eqref{eq:gradient} would be simplified to~$\frac{d\hat{g}(x)}{dx}$. Therefore, \eqref{eq:gradient} is a risk-aware gradient of the GP that models $g$. Note that the derivative of a GP is also a GP, and the calculation of its posterior is given in Sec.~\ref{sec:GP}.

The proposed risk cost function is generic enough to be applicable to several different robots as well as different motion planning algorithms. It can be used in a similar manner as Signed Distance Fields (SDFs) are employed to avoid collision with obstacles. In Sec.~\ref{sec:samplingApproach} we show how the cost map can be incorporated into anytime sampling-based algorithms, which search for optimal trajectories by means of minimizing a functional, e.g. RRT$^\star$ \cite{karaman2011sampling}, BIT$^\star$ \cite{gammell2015batch}, RT-RRT$^\star$ \cite{naderi2015rt} and Online RRT$^\star$ \cite{chandler2017online}. Similarly, in Sec.~\ref{sec:gpmp} we show how to incorporate it into GPMP2 \cite{mukadam2018continuous}, which allows for fast planning and replanning of systems with high degrees-of-freedom.

\section{Sampling-based Approach}\label{sec:samplingApproach}
In order to encode risk assessment into an optimal motion planning algorithm, we propose a cost function that balances the trade-off between trajectory length and level of violation of the constraint $\varphi$. The proposed cost functional $J$ of a trajectory $\varsigma$ is defined as the line integral of the risk cost function $f$ :
\begin{equation}
    J(\varsigma) = \int_C f(x) ds = \int_0^1 f(\varsigma(t)) \|{\varsigma'}(t)\| dt,
    \label{eq:cost}
\end{equation}
where $ds$ is used to denote that the integral is taken along the curve $C$, and $\|{\varsigma'}(t)\|$ is the norm of the directional derivative of the curve. The intuition behind the line integral is that it locally enlarges or shrinks the arc length of a curve according to the value of the cost function, thus of the risk of violating the constraint. 

\begin{rem} \label{lem:lemma1}
    Note that for a trajectory $C$, parameterized by $\varsigma \in \Sigma$, both its arc length $\int_C ds$ and the Euclidean distance between $\varsigma(0)$ and $\varsigma(1)$ are admissible heuristics of the cost functional $J(\varsigma)$. Since $f(x) \ge 0 \;\forall x \in \Ss$, the following is straightforward:
    \begin{equation*}
        J(\varsigma) = \int_C f(x) ds  \ge \int_C ds \ge \|\varsigma(1) - \varsigma(0)\|.
    \end{equation*}
\end{rem}

The proposed sampling-based approach that solves Problem~\ref{prob:prob1} is presented in Algorithm~\ref{alg:base}. The input requirements are i) the specification of the environment $\Ss$ together with its obstacle space $\Ss_{\text{obs}}$ and goal region $\Ss_{\text{goal}}$, ii) an appropriate risk function $r$, safety level $\beta$ and user-defined constraint $\varphi$, iii) an appropriate GP kernel $\kappa$, and iv) a planning budget~$T$ as well as a penalizing parameter $\gamma$ for (\ref{eq:field}). The reason behind the need for a planning budget is because finding the optimal path in finite time is hard, and several motion planners are at most asymptotically optimal; a budget limits the planning algorithm to return its best path when the budget is over (assuming the algorithm is anytime).

The algorithm starts by initializing the data set $\mathcal{D}$ at the agent's initial position $x_{\text{init}}$ with observation  $z(x_{\text{init}})$ and uses it to initialize the GP model (line~\ref{alg:line1}). Then the motion planning algorithm is initialized (line~\ref{alg:line2}), during which the user can define which algorithm will be used, e.g. RRT$^\star$ and BIT$^\star$, and configure its cost function calculation according to Eqs. (\ref{eq:cost}) and (\ref{eq:field}), risk metric $r$, safety level $\beta$ and constraint~$\varphi$. The system is now configured and ready to move to the online part, which runs until the agent has reached a state within the goal region (line \ref{alg:while1}).

Given the current position $x$ of the agent and the GP model $\hat{g}$, the planner searches for a trajectory $\varsigma$ until the computation budged $T$ is exhausted (line~\ref{alg:plan}). The agent then commits to move along the planned trajectory (line~\ref{alg:move}), adding samples of the process to $\mathcal{D}$ as it goes (line~\ref{alg:append}). This is repeated until the agent either arrives at the goal region or replanning is triggered (line~\ref{alg:while2}).

As the framework makes more observations of the unknown process, it builds a more accurate model, and a trajectory that was once deemed optimal might not be so anymore. This is due to the fact that by including the measure of violation of the constraint $\varphi$ inside the cost functional~$J$, the cost map~$f$ changes as new samples are added to~$\mathcal{D}$. The triggering function is a mechanism that monitors the development of the model and detects when replanning is necessary. When such a event occurs, the planner clears its search completely and updates its GP to include the measurements taken while the robot was moving.

\begin{algorithm}[t]
		\small
		\DontPrintSemicolon
		\KwIn{$\Ss$ - environment, $\varphi$ - constraint, $r$ - risk metric, $\beta$ - safety level, $T$ - replanning budget, $x_{\text{init}}$ - initial state, $\Ss_{\text{goal}}$ - goal region, $\kappa$ - GP kernel, $\gamma$}
		\KwOut{sequence of (states, samples) traversed}
		    $\mathcal{D} \leftarrow (x_{\text{init}}, z(x_{\text{init}}))$; \quad
		    $\hat{g} \leftarrow GP(\mathcal{D},\kappa)$ \label{alg:line1}\;
		    $planner.init(\Ss, \varphi, r, \beta, \gamma, \hat{g})$ \label{alg:line2}\;
		    $x \leftarrow x_{\text{init}}$ \;
		    \While{$x \not\in \Ss_{\text{goal}}$}{ \label{alg:while1}
		        $\varsigma \leftarrow planner.plan(x, \Ss_{\text{goal}}, T)$ \label{alg:plan}\;
		        \While{$x \not\in \Ss_{\text{goal}}$ or not(event-trigger())}{\label{alg:while2}
		            $x \leftarrow move(\varsigma)$ \label{alg:move} \;
		            $\hat{g}.\mathcal{D}.append((x, z(x)))$ \label{alg:append}
		            }
		        $planner.clear()$ \label{alg:clear} \;
		        $planner.updateGP(\hat{g})$ \label{alg:update}
		        }
			\Return{$\hat{g}.\mathcal{D}$}
	\caption{{Risk-Aware Sampling-based Planning} \label{alg:base}}
\end{algorithm}

\subsection{Event-triggered replanning}
The need for a replanning event is determined through a comparison between the current risk values on states of the trajectory, and the risk values at the same states given the updated GP. If the updated GP indicates an increase in risk, a replanning event is triggered.  
The motivation behind this design is that the cost functional $J$, definied in Eqs. (\ref{eq:cost}) and (\ref{eq:field}), already takes into account the upper-tail of the probability distribution $\hat{g}(x)$ through the risk metric; however, with new samples of the process included in the GP, the value considered during the previous planning process might not be representative of the current distribution anymore.

Let us denote as $\varsigma_n$ the trajectory found at the $n$-th planning iteration, i.e. $\varsigma_0$ is the initial trajectory found starting at~$x_{\text{init}}$, $\varsigma_1$ is the trajectory found after the first re-planning, and so on. Let us also denote as $\hat{g}_n$ the GP model at iteration~$n$, whose data set is composed of samples taken before that iteration only, and which was used to plan $\varsigma_n$. Lastly, as the robot is moving along $\varsigma_n$ and including more samples in its data set, we denote $\hat{g}^*$ as the most up-to-date GP model.

For a constraint of type $\varphi = \alpha - r_\beta(\hat{g}(x))$, $\forall x \in \varsigma$, i.e. where it is desired to stay below a certain threshold $\alpha \in \Real$, we trigger replanning when
\begin{equation}
    r_\beta(\hat{g}^*(\bar{x})) > r_{\beta'}(\hat{g}_n(\bar{x})) \text{, for } \beta' < \beta,
\end{equation}
and for a certain state $\bar{x} \in \varsigma$, where $\beta'$ is a low safety level that represents the extreme of the upper-tail of the distribution.

\section{Optimization-based Approach} \label{sec:gpmp}
Optimization-based motion planning algorithms have been widely used to refine and post-process trajectories returned by sampling-based planners. One widely-known algorithm is CHOMP \cite{ratliff2009chomp}, a gradient-based optimization technique for refining continuous paths. CHOMP was the baseline for several other approaches, such as STOMP \cite{kalakrishnan2011stomp} and GPMP2~\cite{mukadam2018continuous}. Due to our need for constantly adapting the plan as new information is acquired, the proposed risk cost function is implemented on the incremental version of GPMP2, named iGPMP2. The results can be easily extended to any other optimization-based planner.

The proposed risk-aware iGPMP2 is presented in Alg.~\ref{alg:base2}, which follows steps similar to Alg.~\ref{alg:base}. The main differences come from manipulating the incremental planner, but the overall idea is the same: draw a sample of $g$ and update the GP; optimize trajectory based on the most-recent information; move along the trajectory; take a new sample of $g$; restart the process until the goal has been reached.

We treat the risk-aware cost function in a similar manner to the obstacle cost already used by GPMP2. Recall that GPMP2 optimizes a trajectory by performing inference on a factor graph, which encodes priors over some states. The obstacle factor, a distribution in the exponential family, is 
\begin{equation}
    \mathrm{f}_{obs}(x) = \exp\left\{-\frac{1}{2} \|h(x)\|^2_{\sigma_{obs}}\right\},
\end{equation}
where $h: \Ss \to \Real$ is the obstacle cost function and $\sigma_{obs}$ is a hyperparameter matrix. Thus, in a similar way, we define the risk factor as follows
\begin{equation}
    \mathrm{f}_{risk}(x) = \exp\left\{-\frac{1}{2} \|f(x) - 1\|^2_{\sigma_{risk}}\right\}.
\end{equation}

Alg.~\ref{alg:base2} starts by initializing the GP model with an appropriate kernel and a sample of $g$ at the initial state (line~\ref{alg:line11}), followed by the initialization of the planner (line~\ref{alg:line12}). It is during such an initialization that the user configures the model of the system and the size of the state-space. An initial collision-free trajectory is then searched based only on start and goal states and the obstacle-space (line~\ref{alg:line13}). This initial trajectory is used to initialize the incremental planning algorithm with a good seed (line~\ref{alg:line14}), which speeds-up the process of finding new trajectories in the following steps. The process of initializing the incremental planning algorithm also creates the factor graph that encodes the prior over the trajectory, over the obstacle space, and now also over our risk cost function.

The second part of the algorithm corresponds to the online part, in the sense that it uses the most-recent data about $g$ in order to adapt the current plan according to changes in the environment. Until the goal is reached (line~\ref{alg:while11}), the algorithm iteratively finds a new trajectory (line~\ref{alg:plan1}), moves the robot along the trajectory (line~\ref{alg:move1}), updates the GP model (line~\ref{alg:append1}) and then fixes the factors corresponding to state $x$ (line~\ref{alg:fix}). Fixing such factors will force the incremental planner to not update its values anymore, therefore fixing the trajectory already traversed by the robot.

\begin{algorithm}[t]
		\small
		\DontPrintSemicolon
		\KwIn{$\Ss$ - environment, sdf - signed distance field, $\varphi$ - constraint, $r$ - risk metric, $\beta$ - safety level, $x_{\text{init}}$ - initial state, $x_{\text{goal}}$ - goal state, $\kappa$ - GP kernel, $\gamma$}
		\KwOut{sequence of (states, samples) traversed}
		    $\mathcal{D} \leftarrow (x_{\text{init}}, z(x_{\text{init}}))$; \quad
		    $\hat{g} \leftarrow GP(\mathcal{D},\kappa)$ \label{alg:line11}\;
		    $planner.setup()$ \label{alg:line12}\;
		    $\varsigma \leftarrow planner.initialTraj(x_{\text{init}}, x_{\text{goal}}, sdf)$ \label{alg:line13}\;
		    $iSAM2.init(\varsigma, x_{\text{init}}, x_{\text{goal}}, sdf, \hat{g})$ \label{alg:line14}\;
		    $x \leftarrow x_{\text{init}}$ \;
		    \While{$x \not\in \Ss_{\text{goal}}$}{ \label{alg:while11}
		        $\varsigma \leftarrow iSAM2.plan()$ \label{alg:plan1}\;
		            $x \leftarrow move(\varsigma)$ \label{alg:move1} \;
		            $\hat{g}.\mathcal{D}.append((x, z(x)))$ \label{alg:append1}\;
		            $iSAM2.fixFactors(x)$ \label{alg:fix}
		        }
			\Return{$\hat{g}.\mathcal{D}$}
	\caption{{Risk-Aware iGPMP2} \label{alg:base2}}
\end{algorithm}

\section{Results}

\begin{figure*}
    \centering
    \subfloat[Initial plan\label{fig:draft0}]{\includegraphics[width=0.25\linewidth]{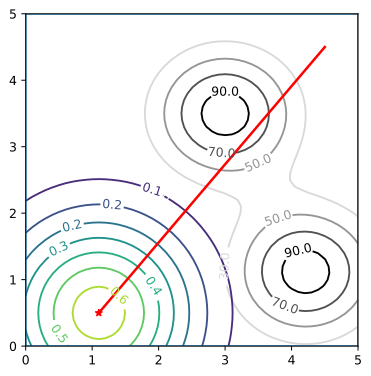}}
    \subfloat[System after 6 replanning triggers\label{fig:draft1}]{\includegraphics[width=0.25\linewidth]{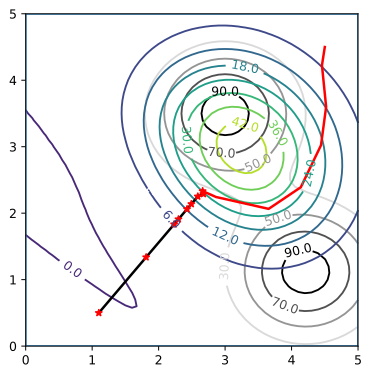}}
    \subfloat[System after 9 replanning triggers\label{fig:draft2}]{\includegraphics[width=0.25\linewidth]{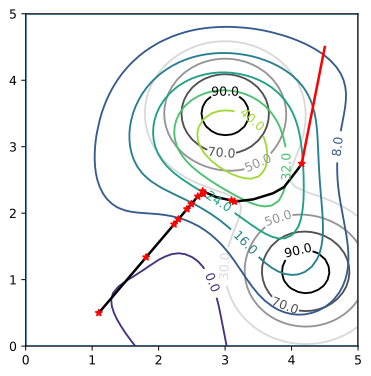}}
    \subfloat[Traversed trajectory\label{fig:draft3}]{\includegraphics[width=0.25\linewidth]{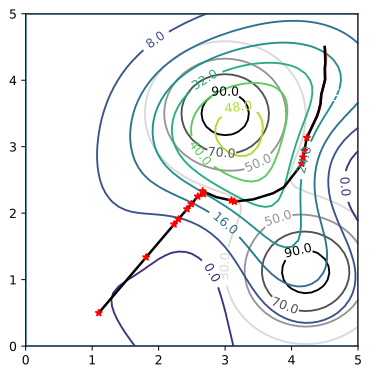}}
    \caption{Snapshots of a point agent using Alg.~\ref{alg:base} to traverse an area, starting from the bottom-left corner, with two sources of hazard. Concentric circles in grayscale are the level sets of the unknown process $g$, while level sets in colors show the current GP belief over $\hat{g}$. The trajectory planned at the current step is presented in red, and trajectory traversed up to that point is in black. Red stars mark locations were the event-based trigger activated the replanning. This simulation used CVaR at safety level $\beta = 0.05$, with replanning budget $T = 3s$ and penalization parameter $\gamma = 0.1$.}
    \label{fig:snaps}
\end{figure*}

\begin{figure*}
    \centering
    \subfloat[Initial plan\label{fig:draft10}]{\includegraphics[width=0.33\linewidth]{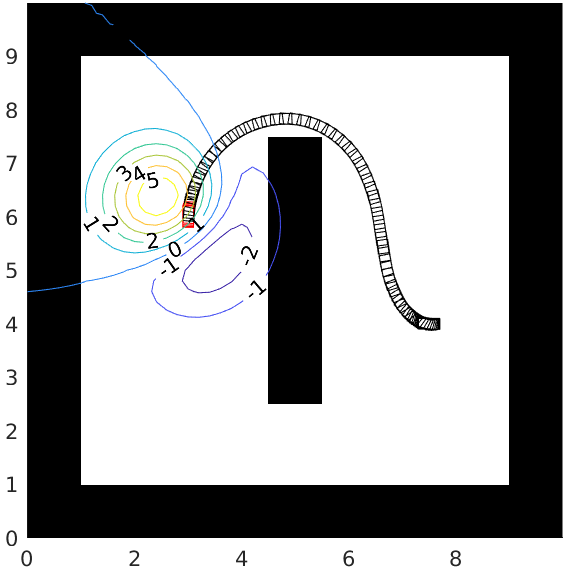}}
    \subfloat[System after navigating a while\label{fig:draft11}]{\includegraphics[width=0.33\linewidth]{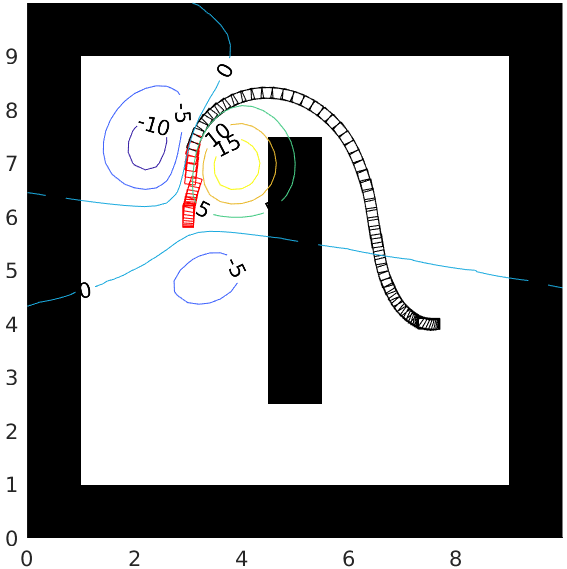}}
    \subfloat[Traversed trajectory\label{fig:draft13}]{\includegraphics[width=0.33\linewidth]{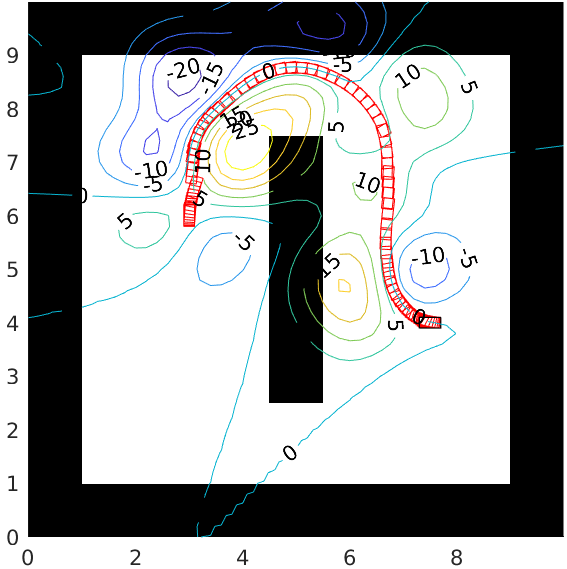}}
    \caption{Snapshots of a 3DOF agent using Alg.~\ref{alg:base2} to traverse an area, starting from the left side, with one source of hazard. The trajectory planned at the current step is presented in black, and trajectory traversed up to that point is in red. This simulation used CVaR at safety level $\beta = 0.2$.}
    \label{fig:snaps1}
\end{figure*}

All the results presented were obtained on an Intel\textsuperscript{\textregistered} Core\textsuperscript{TM} i7-7700HQ CPU with 2.8GHz clock speed and 16GB RAM under Ubuntu 18.04, with a GeForce GTX 1060 GPU. Alg.~\ref{alg:base} was implemented using the open motion planning library (OMPL) \cite{sucan2012ompl} and GPflow \cite{GPflow2017} in Python~3.6. The sampling-based algorithm we used was BIT$^\star$ \cite{gammell2015batch} as it benefits from Remark~\ref{lem:lemma1}. Alg.~\ref{alg:base2} was implemented in C++ using the open-source repository provided by \cite{mukadam2018continuous}.

For the GP model, we used a Squared Exponential kernel with tuned hyperparameters. The variance of the sensor observations were set to $\sigma_n^2 = 0.5$.

\subsection{Sources of hazard}\label{sec:hazard}
In our experiments we assume the existence of at least one source of hazard in the environment, defined as follows. 
As we are dealing with a two-dimensional environment, we denote a position state $x_p := \begin{bmatrix}x_1 & x_2 \end{bmatrix}^T$. We consider the equation of a dynamic function over time as defined by \cite{marchant2014sequential}, which can be used to model the values of a process around a source. Since our problem is not formulated over a dynamic process, we fix time step $\tau$ and use the following equation
\begin{equation}
    g(x_p) = k e^{-\left(\frac{x_1 - c_1 - 1.5\sin{2\pi \tau}}{1.1}\right)} e^{-\left(\frac{x_2 - c_2 - 1.5\cos{2\pi \tau}}{0.9}\right)}.
\end{equation}
Note that by modifying the value of $\tau$ within $[0,1]$, one rotates the source around $\begin{bmatrix}c_1 & c_2 \end{bmatrix}^T \in \Real^2$ along an ellipsoid parameterized by the terms with sine and cosine. Lastly, the values $1.1$ and $0.99$ influence the decay rate in each axis, and $k \in \Real$ is a gain. In all of our scenarios we choose $k = 100$.

\vspace{1cm}
\subsection{Simulation results}
In the first simulation we demonstrate the steps of Alg.~\ref{alg:base} through a sequence of snapshots, Fig.~\ref{fig:snaps}, taken as a point robot traverses an obstacle-free environment while trying to avoid regions where $g$ is greater than 30. The robot is deployed from the bottom-left corner of the environment with the goal of reaching the upper-right corner. Fig.~\ref{fig:draft0} shows the initial configuration of the robot as well as the GP posterior from a single sample at $x_{\text{init}}$, and the trajectory planned at this step. As the robot traverses the environment and takes more samples of the process, the GP posterior changes accordingly, together with the trajectory, as can be seen in Figs.~\ref{fig:draft1} and~\ref{fig:draft2}. Finally, Fig.~\ref{fig:draft3} shows the final trajectory along with the final GP model.

The simulation of Alg.~\ref{alg:base2} follows the same procedure, and is shown in Fig.~\ref{fig:snaps1}. The agent is now a 3-degree-of-freedom car-like robot, the environment contains a central obstacle, and the agent should not be exposed to hazard values above 10. Starting from the left side of the environment, in Fig.~\ref{fig:draft10}, the agent constantly adjusts its trajectory as new samples are taken, with the final trajectory presented in Fig.~\ref{fig:draft13}.

\subsection{Discussion}
According to our problem definition, the mission of the robot is to ultimately go from an initial state to a goal region while minimizing a cost function. We do not aim at performing any type of exploration of the environment to precisely model the unknown process. This can be seen in Fig.~\ref{fig:draft3}, where even though the robot has already completed its mission, the GP posterior still does not closely approximate the process in the entire environment. However, since the robot samples on its own trajectory, its has the highest accuracy where it has travelled through.

We have proposed two applications to the risk-aware cost function described in the paper, and each has presented a few pros and cons. Even though we used BIT$^{\star}$, a guided sampling-based planner that delays the calculation of edge costs in order to save computation effort, there were three major drawbacks: i) edge growth does not exploit any information regarding $g$, such as its gradient, in order to speed up convergence; ii) approximating the cost of an edge is computationally expensive as it relies on constantly calculating GP posteriors; and iii) since the risk cost function is dynamic (it changes as new samples are added to the GP), the graph needs to adapt to it every time replanning is triggered, and doing so is not straightforward.

On the other hand, the incremental approach on iGPMP2 was capable of easily handling the dynamic cost function, quickly converging to a local minima. The drawbacks are more related to the sensitivity of the parameters native to GPMP2.  Such sensitivity oftentimes led to spending too much time tuning the parameters, without the guarantee that the trajectory found is actually dynamically feasible.
Note that, differently from Alg.~\ref{alg:base}, replanning was performed online and at every step in Alg.~\ref{alg:base2}.

We argue that an interesting future work would be to integrate global and local search in a systematic manner, tailored for dynamic cost functions. An example of a similar algorithm but for static cost functions is RABIT$^\star$ \cite{choudhury2016regionally}, which integrates BIT$^\star$ and CHOMP.

\section{Conclusion}

We proposed an approach that allows for online motion planning for robots deployed in partially-known environments, where partially-observable states, modelled by a GP from noisy sensor readings, might pose risk to the agent. We also demonstrated how the proposed risk cost function can be used in both sampling-based and optimization-based planning algorithms. 
Our plans for future work include expanding our results to multi-dimensional processes as well as to testing on real robots in a nuclear inspection task. Furthermore, we also plan to consider more complex trajectory constraints based on STL.

\bibliographystyle{IEEEtran}
\bibliography{literature}

\end{document}